\documentclass[10pt,twocolumn,letterpaper]{article}

\usepackage{iccv}
\usepackage{times}
\usepackage{epsfig}
\usepackage{graphicx}
\usepackage{amsmath}
\usepackage{amssymb}
\usepackage[toc,page]{appendix}
\usepackage{epstopdf}

\DeclareMathOperator*{\argmax}{argmax}

\newenvironment{packed_itemize}{
\vspace{-0.10cm}\begin{itemize}
  \setlength{\itemsep}{1pt}
  \setlength{\parskip}{0pt}
  \setlength{\parsep}{0pt}
}{\end{itemize}}

\newenvironment{packed_enumerate}{
\begin{enumerate}
  \setlength{\itemsep}{1pt}
  \setlength{\parskip}{0pt}
  \setlength{\parsep}{0pt}
}{\end{enumerate}}

\usepackage[breaklinks=true,bookmarks=false]{hyperref}

\iccvfinalcopy 

\def\iccvPaperID{****} 

\ificcvfinal\fi
\begin{document}

\title{Be Your Own Prada: Fashion Synthesis with Structural Coherence\thanks{This is the updated version of our original paper appeared in ICCV 2017 proceedings.}}

\author{Shizhan Zhu$^{1}$ \quad Sanja Fidler$^{2,3}$ \quad Raquel Urtasun$^{2,3,4}$  \\ \quad Dahua Lin$^{1}$  \quad Chen Change Loy$^{1}$  \\[1mm]
$^1$Department of Information Engineering, The Chinese University of Hong Kong \\ $^2$University of Toronto, \hspace{3mm}$^3$Vector Institute, \hspace{3mm}$^4$Uber Advanced Technologies Group\\
{\tt\small \{szzhu, dhlin, ccloy\}@ie.cuhk.edu.hk, \{fidler, urtasun\}@cs.toronto.edu}
}

\maketitle

\begin{abstract}
\vspace{-0.2cm}
We present a novel and effective approach for generating new clothing on a wearer through generative adversarial learning.
Given an input image of a person and a sentence describing a different outfit, our model ``redresses'' the person as desired, while at the same time keeping the wearer and her/his pose unchanged.
Generating new outfits with precise regions conforming to a language description while retaining wearer's body structure is a new challenging task.
Existing generative adversarial networks are not ideal in ensuring global coherence of structure given both the input photograph and language description as conditions.
We address this challenge by decomposing the complex generative process into two conditional stages. In the first stage, we
generate a plausible semantic segmentation map that obeys the wearer's pose as a latent spatial arrangement. An effective spatial constraint is formulated to guide the generation of this semantic segmentation map.
In the second stage, a generative model with a newly proposed compositional mapping layer is used to render the final image with precise regions and textures conditioned on this map.
We extended the DeepFashion dataset~\cite{liu2016deepfashion} by collecting sentence descriptions for 79K images. We demonstrate the effectiveness of our approach through both quantitative and qualitative evaluations. A user study is also conducted.
The codes and the data are available at \url{http://mmlab.ie.cuhk.edu.hk/projects/FashionGAN/}.
\end{abstract}

\vspace{-0.4cm}

\section{Introduction}
\label{sec:introduction}

\begin{figure}[t]
\begin{center}
\includegraphics[width=\linewidth]{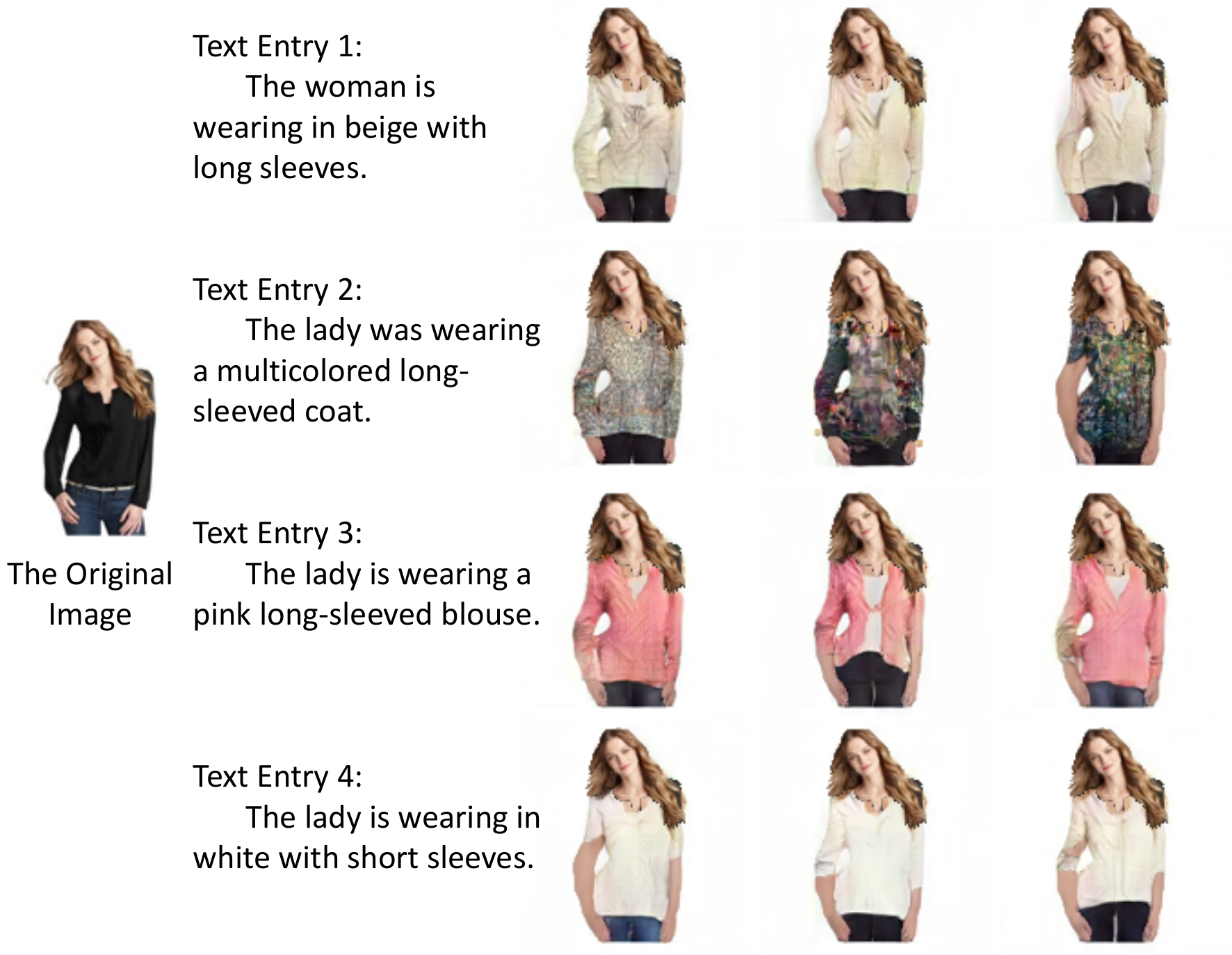}
\caption{Given an original wearer's input photo (left) and different textual
descriptions (second column),
our model generates new outfits onto the photograph (right three columns)
while preserving the pose and body shape of the wearer.}
\vskip -0.7cm
\label{fig:fig1}
\end{center}
\end{figure}

Imagine that you could be your own fashion
designer, and be able to seamlessly transform your current outfit in the photo
into a completely new one, by simply describing it in words (Figure~\ref{fig:fig1}). In just minutes you could design and ``try on'' hundreds
of different shirts, dresses, or even styles, allowing you to easily discover
what you look good in.
%
The goal of this paper is to develop a method that can generate new outfits onto
existing photos, in a way that preserves \emph{structural coherence} from multiple perspectives:
\begin{packed_enumerate}
\item Retaining the body shape and pose of the wearer,
\item Producing regions and the associated textures that conform to the language description, and
\item Enforcing coherent visibility of body parts.
\end{packed_enumerate}
%

Meeting \emph{all} these requirements at the same time is a very challenging
task. First, the input image is the only source from which we can mine for the
body shape information. With only a single view of the wearer, it is nontrivial
to recover the body shape accurately. Moreover, we do not want the shape of the
generated outfit to be limited by the original garments of the wearer. For
example, replacing the original long-sleeve shirt with a short-sleeve garment
would require the model to hallucinate the person's arms and skin.

Conventional 2D non-parametric methods or 3D graphics approaches meet the first requirement
through structural constraints derived from human priors.
They can be in the form of accurate physical measurements
(\eg~height, waist, hip, arm length) to create 3D virtual
bodies~\cite{protopsaltou2002body}; manual manipulations of sliders such as
height, weight and waist girth \cite{zhou2010parametric}; or indication of joint
positions and a rough sketch outlining the human body
silhouette~\cite{yang2016detailed}.
All these methods require explicit human interventions at test time,
which would limit their applicability in practical settings.
In addition, as these methods provide no obvious ways to incorporate textual
descriptions to condition the synthesis process, it is non-trivial to fulfil the second requirement with existing methods.
Lastly, they do not meet the third requirement as they do not support hallucination of the missing parts.

Generative Adversarial Networks (GAN)~\cite{goodfellow2014generative}
is an appealing alternative to conventional methods. In previous work,
DCGAN~\cite{radford2015unsupervised}, a GAN formulation combined with convolutional networks,
has been shown to be an effective model to produce realistic images.
Moreover, it allows for an end-to-end embedding of textual descriptions to
condition the image generation.
The task of clothing generation presents two significant challenges
which are difficult to address with the standard DCGAN. First, it directly targets the pixel values
and provides no mechanism to enforce structural coherence \wrt to the input.
Second, it tends to average out the pixels~\cite{salimans2016improved},
thus resulting in various artifacts, \eg~blurry boundaries,
as shown by our experiments.

To tackle this problem, we propose an effective two-stage GAN framework that
generates shape and textures in different stages.
In the first stage, we aim to generate a plausible human segmentation
map that specifies the regions for body parts
and the upper-body garment. This stage is responsible for preserving
the body shape and ensuring the coherent visibility of parts based on the
description.
In the second stage, the generator takes both the produced segmentation
map and the textual description as conditions, and renders the region-specific
texture onto the photograph.

To ensure the coherence in structure of the synthesized image with respect to the input image
(\ie~preserving the body shape and pose of the wearer), we present an effective spatial constraint
that can be derived from the input photograph.
We formulate it carefully so that it does not contradict to
the textual description when both of them are used to condition the first-stage GAN.
In addition, we also introduce a new compositional mapping layer into the
second-stage GAN to enforce region-specific texture rendering guided by the segmentation map.
In contrast to existing GANs that perform non-compositional synthesis,
the new mapping layer is capable of generating more coherent visibility of
body parts with image region-specific textures.

To train our model, we extend the DeepFashion dataset~\cite{liu2016deepfashion}
by annotating a subset of $79K$ upper-body images with sentence descriptions
and human body annotations\footnote{The data and code can be found at \url{http://mmlab.ie.cuhk.edu.hk/projects/FashionGAN/}.}. 
Extensive quantitative and qualitative comparisons are performed against existing
GAN baselines and 2D non-parametric approaches. We also  conduct a user study in order to obtain an
objective evaluation on both the shape and image generation results.


\section{Related Work}
\label{sec:related_work}

%

Generative Adversarial Networks (GAN)~\cite{goodfellow2014generative} have shown
impressive results generating new
images, \eg~faces~\cite{radford2015unsupervised}, indoor
scenes~\cite{wang2016generative}, fine-grained objects like
birds~\cite{reed2016generative}, or clothes~\cite{yoo2016pixel}.
%
Training GANs based on conditions incorporates further information to guide the
generation process.
Existing works have explored various conditions, from category
labels~\cite{nguyen2016synthesizing}, text~\cite{reed2016generative}
to an encoded feature vector~\cite{yoo2016pixel}.
Different from the studies above, our study aims at generating the target
by using the spatial configuration of the input images as a condition.
The spatial configuration is carefully formulated so that it is agnostic to the clothing worn in the original image, and only captures information about the user's body.



There exist several studies to transfer an input image to a new one.
Ledig \etal~\cite{ledig2017photo} apply the GAN framework to
super-resolve a low-resolution image.
Zhu \etal~\cite{isola2017image} use a conditional GAN to transfer
across the image domains, \eg~from edge maps to real images, or
from daytime images to night-time.
Isola \etal~\cite{isola2017image} change the viewing angle of an
existing object.
Johnson \etal~\cite{johnson2016perceptual} apply GANs to neural
style transfer.
All these studies share a common feature - the image is transformed
\emph{globally} on the texture level but is not region-specific.
In this study, we explore a new compositional mapping method that allows
region-specific texture generation, which provides richer textures for different
body regions.

There are several recent studies that explore improved image generation by stacking GANs.
Our work is somewhat similar in spirit to~\cite{wang2016generative, zhang2017stackgan} -- our idea is to have the first stage to create the basic composition, and the second stage to add the necessary refinements to the image generated in the first stage.
However,  the proposed FashionGAN differs from S$^2$GAN~\cite{wang2016generative} in that the latter aims at synthesizing a
surface map from a random vector in its first stage. In contrast, our goal is to generate a plausible mask whose structure conforms to a given photograph and language description, which requires us to design additional spatial constraints and design coding as conditions. Furthermore, these two conditions should not contradict themselves.
Similarly, our work requires additional constraints which are not explored in~\cite{zhang2017stackgan}. Compositional mapping is not explored in the aforementioned studies as well.


Yo \etal~\cite{yoo2016pixel} propose an image-conditional image generation model to perform domain transfer, \eg, generating a piece of clothing from an input image of a dressed person.
Our work differs in that we aim at changing the outfit of a person into a newly designed one
based on a textual description.
Rendering new outfits onto photographs with unconstrained human poses bring additional
difficulties in comparison with work that generates pieces of clothing in a fixed view-angle as in
\cite{yoo2016pixel}.



\begin{figure*}[t]
\begin{center}
\includegraphics[width=\linewidth]{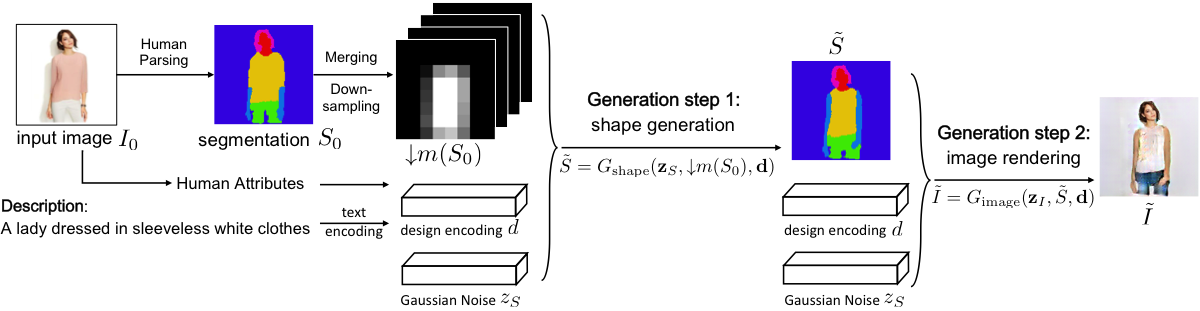}
\caption{\textbf{Proposed framework}. Given an input photograph of a person and a sentence description of a new desired outfit, our model first generates a segmentation map $\tilde{S}$ using the generator from the first GAN. We then render the new image with another GAN, with the guidance from the segmentation map generated in the previous step. At test time, we obtain the final rendered image with a forward pass through the two GAN networks.}
\label{fig:overview}
\end{center}
\vspace{-0.3cm}
\end{figure*}

\section{Methodology}
\label{sec:methodology}

Our framework is inspired by the generative adversarial network (GAN) proposed by Goodfellow~\etal~\cite{goodfellow2014generative}.
We first provide a concise review of GAN, and then introduce our outfit generation framework.
Generative Adversarial Network~\cite{goodfellow2014generative} has shown a powerful capability in generating realistic natural images. A typical GAN contains a generator $G$ and a discriminator $D$.
They are jointly trained with a learning objective given below:
\begin{equation}
\label{eqn:original_gan_loss}
\min_G \max_D \ \mathbb{E}_{I \sim p_\mathrm{data}} [\log D(I)] +
\mathbb{E}_{\mathbf{z} \sim p_z} [\log (1-D(G(\mathbf{z})))].
\end{equation}
Here, $\mathbf{z}$ is a random or encoded vector,
$p_\mathrm{data}$ is the empirical distribution of training images,
and $p_z$ is the prior distribution of $\mathbf{z}$.
It was proven in~\cite{goodfellow2014generative} that when it reaches
the maximum, the distribution of $G(\mathbf{z})$ would converge to $p_{data}$,
where the discriminator cannot distinguish the images $I \sim p_{data}$ from the generated ones.


\subsection{Overview of FashionGAN}
\label{subsec:fashion_generation}

We define the problem as follows. We assume we have the original image of a wearer and a sentence description of the new outfit. An example of a description we envision is
{\em ``a white blouse with long sleeves but without a collar''}.
Our goal is to produce a new image of the user wearing the desired outfit.

Our method requires training data in order to learn the mapping from one photo to the other given the description.
We do not assume paired data where the same user is required to wear two outfits (current, and the described target outfit). Instead, we only require one photo per user where each photo has a sentence description of the outfit. Such data is much easier to collect (Sec.~\ref{subsec:details}).

Since in our scenario we only have one (described) image per user, this image serves as both the input and the target during training.
Thus, rather than working directly with the original image $I_0$, we extract the person's segmentation map, $S_0$, which contains pixel-wise class labels such as \emph{hair}, \emph{face}, \emph{upper-clothes}, \emph{pants}/\emph{shorts}, etc. 
The segmentation map is thus capturing the shape of the wearer's body and parts, but not their appearance.

To capture further information about the wearer, we extract a vector of binary attributes, $\mathbf{a}$, from the person's face, body and other physical characteristics. Examples of attributes include gender, long/short hair, wearing/not wearing sunglasses and wearing/not wearing hat. The attribute vector may additionally capture the mean RGB values of skin color, as well as the aspect ratio of the person, representing coarse body size. These are the properties that our final generated image should ideally preserve. Details of how we extract this information are given in Sec.~\ref{subsec:details}.

We represent the description as a vector $\mathbf{v}$ using an existing text encoder (details in Sec.~\ref{subsec:details}). Our problem is then formalized as follows. Given  $\mathbf{d} = (\mathbf{a},\mathbf{v})$, which we call the \emph{design coding}, and the human segmentation map $S_0$, our goal is to synthesize a new high-quality image $\tilde I$ of the wearer matching the requirements provided in the description, while at the same time preserving the wearer's pose and body shape. Note that during training, $\tilde I=I_0$.

As shown in Fig.~\ref{fig:overview}, we decompose the overall generative process into
two relatively easier stages, namely the human segmentation (shape) generation (corresponding to the desired/target outfit) and texture rendering. This decomposition can be expressed as follows:
\begin{align}
&\tilde{S} \leftarrow G_\mathrm{shape}(\mathbf{z}_S,\downarrow \hspace{-0.1cm} m(S_0),\mathbf{d}), \; \; \label{eqn:s1}\\
&\tilde{I} \leftarrow G_\mathrm{image}(\mathbf{z}_I,\tilde{S},\mathbf{d}). \label{eqn:s2}
\end{align}
Here, $G_\mathrm{shape}$ and $G_\mathrm{image}$ are two separate generators.

More precisely, in our first stage (Eq.~\eqref{eqn:s1}), we first generate a human segmentation map $\tilde{S}$ by taking the original segmentation map $S_0$ and the design coding $\mathbf{d}$ into account.
Here $\downarrow \hspace{-0.1cm} m(S_0)$ is a low-resolution representation of $S_0$, serving as the \textbf{spatial constraint} to ensure structural coherence of the generated map $\tilde{S}$ to the body shape and pose of the wearer.
In the second stage (Eq.~\eqref{eqn:s2}), we use the generated segmentation map $\tilde{S}$ produced by the first generator, as well as the design coding $\mathbf{d}$, to render the garments for redressing the wearer.
The texture for each semantic part is generated in different specialized channels, which are then combined according to the segmentation map $\tilde{S}$ to form the final rendered image. We call this process a \textbf{compositional mapping}.
This newly introduced mapping is useful for generating high-quality texture details within specific regions.

We provide details of the two generators in our framework in Sec.~\ref{subsec:s1} and Sec.~\ref{subsec:s2}, respectively.

\subsection{Segmentation Map Generation ($G_\mathrm{shape}$)}
\label{subsec:s1}

Our first generator $G_\mathrm{shape}$ aims to generate the semantic segmentation map $\tilde{S}$ by conditioning on the spatial constraint $\downarrow \hspace{-0.1cm} m(S_0)$, the design coding $\mathbf{d}\in \mathbb{R}^{D}$, and the Gaussian noise $\mathbf{z}_S \in \mathbb{R}^{100}$. We now provide more details about this model.
%
%
To be specific, assume that the original image is of height $m$ and width $n$, \ie, $I_0 \in \mathbb{R}^{m\times n \times 3}$. We represent the segmentation map $S_0$ of the original image using a pixel-wise one-hot encoding, \ie, $S_0 \in \{0,1\}^{m \times n \times L}$, where $L$ is the total number of labels. In our implementation, we use $L=7$ corresponding to \emph{background}, \emph{hair}, \emph{face}, \emph{upper-clothes}, \emph{pants}/\emph{shorts}, \emph{legs}, and \emph{arms}.

\begin{figure}[t]
\begin{center}
\includegraphics[width=\linewidth]{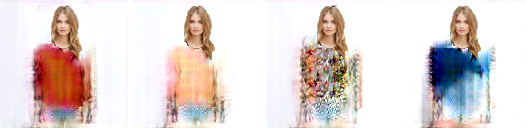}
\caption{This figure motivates the use of $\downarrow \hspace{-0.1cm} m(S_0)$ as a spatial constraint in the first-stage of FashionGAN. Using the high-resolution segmentation map $S_0$ will result in artifacts in the final generated image, when the segmentation maps convey semantic meanings that differ from the textual description. See Sec.~\ref{subsec:s1} for details.}
\label{fig:f1}
\end{center}
\vspace{-0.4cm}
\end{figure}

\vspace{0.1cm}
\noindent
\textbf{Spatial Constraint $\downarrow \hspace{-0.3cm} m(S_0)$}.
We merge and down-sample the original segmentation map $S_0$ into $\downarrow \hspace{-0.1cm} m(S_0) \in [0,1]^{m' \times n' \times L'}$ ($L' < L$ in our implementation), as a conditioning variable to $G_\mathrm{shape}$. In particular, we use $L'=4$ categories: \emph{background}, \emph{hair}, \emph{face}, and \emph{rest}. This essentially maps all the clothing pixels into a generic \emph{rest}
(or \emph{body}) class. Thus, $\downarrow \hspace{-0.1cm} m(S_0)$ is agnostic of the clothing worn in the original image, and only captures information about the user's body.
%
This spatial constraint plays an important role in preserving  structural coherence of the generated shape $\tilde{S}$, while still allowing variability in the generative process.

%
%
We use a down-sampled version of $S_0$ as a constraint so as to weaken the correlation between the two conditions $S_0$ and $\mathbf{d}$, which can contradict each other. Specifically, while $S_0$ keeps the complete information of the wearer's body shape, its internal partitioning of regions do not necessarily agree with the specifications conveyed in the design coding $\mathbf{d}$.
If we were to directly feed the high-resolution segmentation map of the original image into the model, strong artifacts would appear when the textual description contradicts with the segmentation map,
\eg,~the model simultaneously receives the text description ``to generate a long dress'' and the segmentation map that indicates short upper clothes. Figure~\ref{fig:f1} shows such failure cases.

\begin{figure}[t]
\begin{center}
\includegraphics[width=\linewidth]{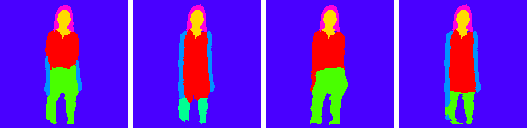}
\caption{Conditioning on the same input image, the $G_{shape}$ generates different human segmentation maps based on different design coding $\mathbf{d}$ and a random vector $\mathbf{z}$. We can observe clear shape differences in sleeves, the length of the upper-clothes, and the labels assigned to legs/pants across the different samples.}
\label{fig:s1}
\end{center}
\vskip -0.4cm
\end{figure}

\vspace{0.1cm}
\noindent
\textbf{Shape Generation}.
We want our $G_\mathrm{shape}$ to output a new human segmentation map $\tilde{S} \in [0,1]^{m \times n \times L}$. This output should ideally have attributes consistent with the design coding $\mathbf{d}$, while at the same time, the generated human shape should conform to the human pose as encoded in the original $S_0$.
The generated segmentation map $\tilde{S}$ should differ from the original human shape $S_0$ with new variations introduced by the design coding $\mathbf{d}$ and noise $\mathbf{z}_S$.
Figure~\ref{fig:s1} illustrates an example of the generated segmentation map. We observe that while the length of the sleeve and upper-clothes vary in different generated samples, the human pose and body shape remain consistent.

To produce the segmentation map $\tilde{S}$, we employ a GAN to learn the generator $G_\mathrm{shape}$.
Both the generator and discriminator comprise of convolution / deconvolution layers with batch normalization and non-linear operations.
Note that different from most of the existing GANs for image generation, the shape map $\tilde{S}$ we are generating in this step is governed by additional constraints -- each pixel in the map has a probabilistic simplex constraint, i.e. $\tilde{S}_{ij} \in \Delta^L, 1 \leq i \leq m, 1 \leq j \leq n$.
We use the Softmax activation function on each pixel at the end of the generator, so that the generated fake shape map is comparable with the real segmentation map.
We observe that the GAN framework can also learn well in this scenario.
Please refer to \textit{suppl. material} for a detailed description of the network structure.

\subsection{Texture Rendering ($G_\mathrm{image}$)}
\label{subsec:s2}

Having obtained the human segmentation map $\tilde{S} \in \mathbb{R}^{m \times n \times L}$ from the generator $G_\mathrm{shape}$, we now use this map along with the design coding vector $\mathbf{d} \in \mathbb{R}^{D}$ to render the final image $\tilde{I} \in \mathbb{R}^{m \times n \times 3}$ using the second-stage $G_\mathrm{image}$.

\vspace{0.1cm}
\noindent
\textbf{Compositional Mapping}.
Conventional GANs generate an image without enforcing region-specific texture rendering.
In the proposed FashionGAN, we propose a new compositional mapping layer that generates the image with the guidance of the segmentation map.
In comparison to non-compositional counterparts, the new mapping layer helps to generate textures more coherent to each region and maintain visibility of body parts.

Formally, we train a specific channel in $G_\mathrm{image}$ for each category $l$, where $1 \leq l \leq L$, and $L$ is the total number of labels in the segmentation map.
We denote the set of pixels that belong to category $l$ as $P_l$, and form our final generated image $\tilde{I}$ as a collection of pixels $(\tilde{I})_p$ indexed by $p$,
\begin{equation}
(\tilde{I})_p = \sum\nolimits_{l=1}^L \mathbf{1}_{p \in P_l} \cdot (\tilde{I}_l)_p,
\label{eqn:comp}
\end{equation}
where $p$ is the index of the pixel and $\tilde{I}_l$ is the specific channel for the $l$-th category. Here $\mathbf{1}_{(\cdot)}$ is an indicator function.
%


\vspace{0.1cm}
\noindent
\textbf{Image Generation}.
Similar to the networks in Sec.~\ref{subsec:s1}, the generator and discriminator in this step are also composed of convolution / deconvolution layers with batch normalization and non-linear operations.
Instead of assigning a Tanh activation function at the end of the network as most of GAN architectures do, we put this activation before the region-specific rendering layer $\tilde{I}_l$.
This is important for achieving a stable combination of all the channels generated by the network.
Please refer to \textit{supplementary material} for a detailed description of the network structure.

\subsection{Training}
\label{subsec:training}

Our two GANs are trained separately due to the non-differentiable $\argmax$ operation between the two steps.
%
%
The training process needs one fashion image $I_0$ for each person in the training set, along with the textual description (represented by its designing coding $\mathbf{d}$) and the segmentation map $S_0$.
%
%
In our first GAN, we derive the tuple $\{\downarrow \hspace{-0.1cm} m(S_0), \mathbf{d}, S_0\}$ from each training sample and train the networks, following the typical conditional GAN training procedure. In our second GAN, we derive the tuple $\{\downarrow \hspace{-0.1cm} m(S_0), \mathbf{d}, I_0\}$ from each training sample for training. We use the Adam optimizer~\cite{kingma2014adam} in training.
Discriminative networks only appear in the training phase. Similar to \cite{reed2016generative}, we provide the conditions (design coding, segmentation maps) to the discriminative networks to enforce consistency between the conditions and the generated results.

\subsection{Implementation Details and Dataset}
\label{subsec:details}

The dimensionality of the design coding $\mathbf{d}$ is $D=50$.
Ten dimensions in $\mathbf{d}$ serve as the human attributes. We represent the binary attributes of gender, long/short hair, w/o sunglasses, w/o hat with one dimension each. We extract the medium value of the R, G, B as well as the Y (gray) channel among the skin region, a total of four dimensions, to represent the skin color. We use the height and width of the given person to represent the size as well as the aspect ratio.
The remaining 40 dimensions are the encoded text.
We follow~\cite{reed2016generative} to construct the text encoder, which can be jointly tuned in each of the GANs in our framework.
The resolution of our output image $\tilde{I}$ is 128$\times$128 (i.e. $m=n=128$).

%
%
We perform bicubic down-sampling to get $\downarrow \hspace{-0.1cm} m(S_0)$, with a size of 8$\times$8 (i.e. $m'=n'=8$).
 We keep the hair and face regions in our merged maps avoiding the need for the generator to generate the exact face as the original wearer (we replace the generated hair/face region with the original image $I_0$). It is hard and not necessary in practice.
%
%
%
%
%

To train our framework 
we extended the publicly available DeepFashion dataset~\cite{liu2016deepfashion} with richer annotations (captions and segmentation maps).
In particular, we selected a subset of 78,979 images from the DeepFashion attribute dataset, in which the person is facing toward the camera, and the background of the image is not severely cluttered.
Training our algorithm requires segmentation maps and captions for each image.
We manually annotated one sentence per photo, describing only the visual facts (\eg,~the color, texture of the clothes or the length of the sleeves), avoiding any subjective assessments.
For segmentation, we first applied a semantic segmentation method (VGG model fine-tuned on the ATR dataset~\cite{liang2015human}) to all the images, and then manually checked correctness.
We manually relabeled the incorrectly segmented samples with GrabCut~\cite{rother2004grabcut}.

\section{Experiments}
\label{sec:experiments}

We verify the effectiveness of FashionGAN through both quantitative and qualitative evaluations. Given the subjective nature of fashion synthesis, we also conduct a blind user study to compare our method with 2D nonparametric based method and other GAN baselines.

\noindent
\textbf{Benchmark.} We randomly split the whole dataset (78,979 images) into a disjoint training set (70,000 images) and test set (8,979 images).
%
%
All the results shown in this section are drawn from the test set.
A test sample is composed of a given (original) image and a sentence description serving as the redressing condition. 
%

\noindent
\textbf{Baselines.} As our problem requires the model to generate a new image by keeping the person's pose, many existing unconditional GAN-based approaches (\eg,~DCGAN~\cite{radford2015unsupervised}) are not directly applicable to our task.
Instead, we use the conditional variants to serve as the baseline approach in our evaluation. We compare with several baselines as follows:

\begin{table*}[t]
\centering
\caption{Evaluation results of detecting five structure-relevant attributes from synthesized images. Average precision is reported.} \vspace{0.1cm}
\label{tab:attr}
\small{
\begin{tabular}{ccccccc}
\hline
Images from            & Has T-Shirt   & Has Long Sleeves & Has Shorts    & Has Jeans     & Has Long Pants & mAP           \\ \hline \hline
Original (Upper-bound) & 77.6          & 88.2             & 90.4          & 86.5          & 91.2           & 86.8          \\ \hline
One-Step-8-7           & 56.5          & 73.4             & 79.6          & 73.5          & 79.1           & 72.4          \\
One-Step-8-4           & 57.1          & 75.0             & 80.1          & 74.3          & 79.8           & 73.3          \\
Non-Compositional      & 54.7          & 79.2             & 82.3          & 72.8          & 84.4           & 74.7          \\
Non-Deep & 56.1 & 69.4 & 75.7 & 74.3 & 76.5 & 71.8 \\
\textbf{FashionGAN}          & \textbf{63.2} & \textbf{86.9}    & \textbf{90.0} & \textbf{82.1} & \textbf{90.7}  & \textbf{82.6} \\ \hline
\end{tabular}
}
\end{table*}


\noindent
\textbf{(1) One-step GAN}: To demonstrate the effectiveness of the proposed two-step framework, we implemented a conditional GAN to directly generate the final image in one step, \ie, $\tilde{I} = G_\text{direct}(\mathbf{z}, S_0, \mathbf{d})$. We refer to this type of baseline as \textit{One-Step}.
Since we aim to generate a new outfit that is consistent with the wearer's pose in the original photo, the one-step baseline also requires similar spatial priors.
Recall that we need to avoid contradiction between conditions from the text description and segmentation (see Sec.~\ref{subsec:fashion_generation}).
Hence, for a fair comparison between our proposed approach and this baseline, we feed in the down-sampled version of the ground-truth segmentation map.
We further divide this type of baseline into two different settings based on the way we use the shape prior $S_0$:
\begin{packed_itemize}
\item \textit{One-Step-8-7}: We use the down-sampled but not merged segmentation map (8$\times$8$\times$7) as the prior;
\item \textit{One-Step-8-4}: We use the down-sampled merged segmentation map (8$\times$8$\times$4) as the prior (the same setting we used in our first stage GAN).
\end{packed_itemize}
The architecture of the generator and discriminator used in these baselines are consistent to those used in our proposed method, i.e., containing 6 deconvolution and convolution layers in both the generator and discriminator.

\vspace{0.1cm}
\noindent
\textbf{(2) Non-Compositional}: To demonstrate the effectiveness of the segmentation guidance, we build a baseline that generates an image as a whole, \ie, without using Eq.~\eqref{eqn:comp}. In this baseline, we use two generative stages as in our proposed framework. In addition, the first stage generator of this baseline is still conditioned on the spatial constraint to ensure structure coherence to the wearer's pose.
%
%

\begin{figure}[t]
\begin{center}
\includegraphics[width=\linewidth]{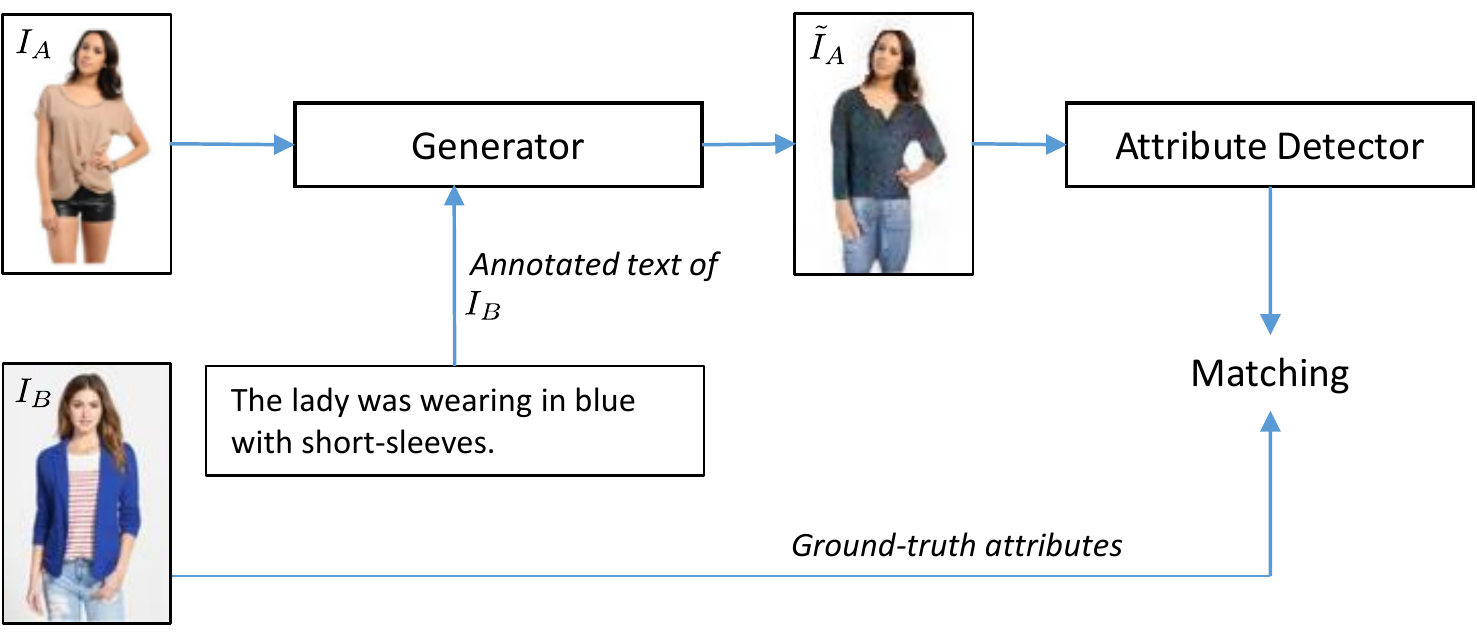}
\caption{The experimental procedure to quantitively verify the attribute and structure consistency of a generated image.}
\label{fig:attribute_consistency_exp}
\end{center}
\vskip -0.5cm
\end{figure}


\subsection{Quantitative Evaluation}
\label{subsec:quantitative_evaluation}

A well-generated fashion image should faithfully produce regions and the associated textures that conform to the language description.
This requirement can be assessed by examining if the desired outfit attributes are well captured by the generated image. 
In this section, we conduct a quantitative evaluation of our approach to verify the capability of FashionGAN in preserving attribute and structural coherence with the input text.

%
We selected a few representative attributes from DeepFashion, namely,  `Has T-Shirt', `Has Long Sleeves', `Has Shorts', `Has Jeans', `Has Long Pants'. These attributes are all \textit{structure-relevant}. A generative model that is poor in maintaining structural coherence will perform poorly on these attributes.
Specifically, we performed the following experiment, as illustrated in Fig.~\ref{fig:attribute_consistency_exp}.
(1) For each test image $I_A$, we used a sentence of another randomly selected image $I_B$ as the text input. The same image-text pairs were kept for all baselines for a fair comparison.
(2) We used the image-text pair as input and generated a new image $\tilde{I}_A$ using a generative model.
(3) We used an external attribute detector\footnote{We fine-tuned the R*CNN model~\cite{gkioxari2015contextual} on our training set to serve as our attribute detector.} to predict the attributes on $\tilde{I}_A$.
(4) Attribute prediction accuracy was computed by verifying the predictions on $\tilde{I}_A$ against ground-truth attributes on $I_B$.

Table~\ref{tab:attr} summarizes the attribute prediction results.
It can be observed that attribute predictions yielded by FashionGAN are more accurate than other baselines. In particular, our approach outperforms one-step GANs that come without the intermediate shape generation, and two-stage GAN that does not perform compositional mapping. Moreover, the performance of FashionGAN is close to the upper-bound, which was provided by applying the attribute detector on image $I_B$, where the text input originated from.
The results suggest the superiority of FashionGAN in generating fashion images with structure coherence.


\subsection{Qualitative Evaluation}
\label{subsec:qualitative_evaluation}

\begin{figure*}[t]
\begin{center}
\includegraphics[width=0.83\linewidth]{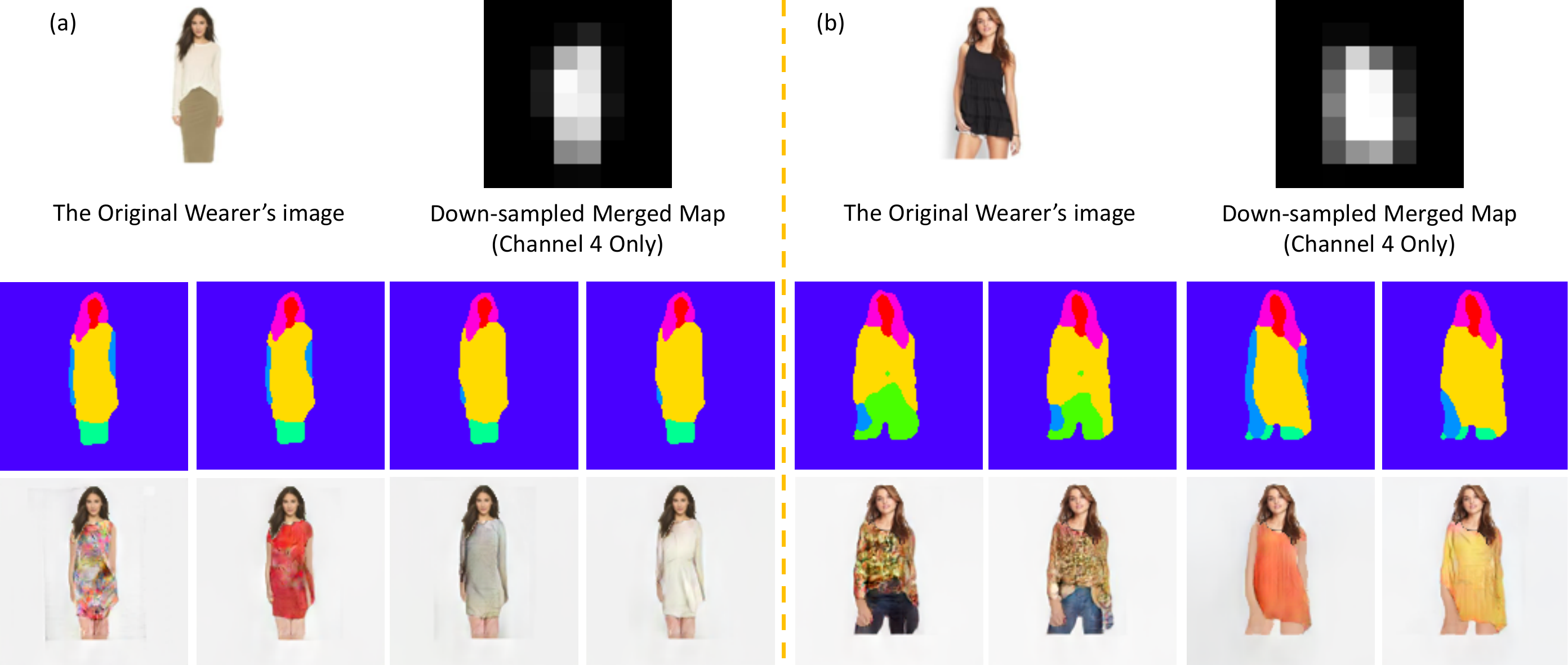}
\caption{\textbf{Conditioning on the same wearer}. A step-by-step visualization of the generation process on two original images (here the sentence descriptions are different for all the 8 samples). Based on the original image and the down-sampled merged segmentation map (top), we first generate a plausible human shape (the middle row), and then use this as priors to guide the texture synthesis in the succeeding step (the bottom row).
}
\vspace{-0.15in}
\label{fig:e1}
\end{center}
\end{figure*}

\noindent
\textbf{Conditioning on the Same Wearer}. Given an image, we visualize the output of FashionGAN with different sentence descriptions. We show all the intermediate results and final rendering step-by-step in Fig.~\ref{fig:e1}, showcasing our generation process. 
A plausible segmentation map is generated first, and one can notice the variation in shape (\eg, the length of the sleeve). The image generated in the second step has consistent shape with the shape generated in the first step. The generated samples demonstrate variations in textures and colors, while the body shape and pose of the wearer are retained.

\begin{figure}[t]
\begin{center}
\includegraphics[width=\linewidth]{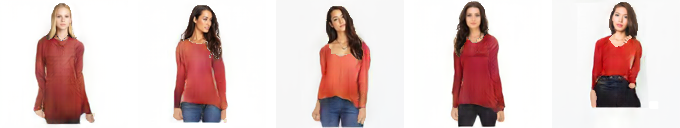}
\caption{\textbf{Conditioning on the same sentence description}. Samples conditioned on the same sentence ``A lady dressed in red clothes with long sleeves''.}
\vspace{-0.25in}
\label{fig:e3}
\end{center}
\end{figure}

\noindent
\textbf{Conditioning on the Same Description}. In this experiment, we choose photos of different wearers but use the same description to redress them. We provide results in Fig.~\ref{fig:e3}. Regardless of the variations in the human body shapes and poses, our model consistently generates output that respects the provided sentence, further showing the capability of FashionGAN in retaining structural coherence.

\begin{figure*}[t]
\begin{center}
\includegraphics[width=\linewidth]{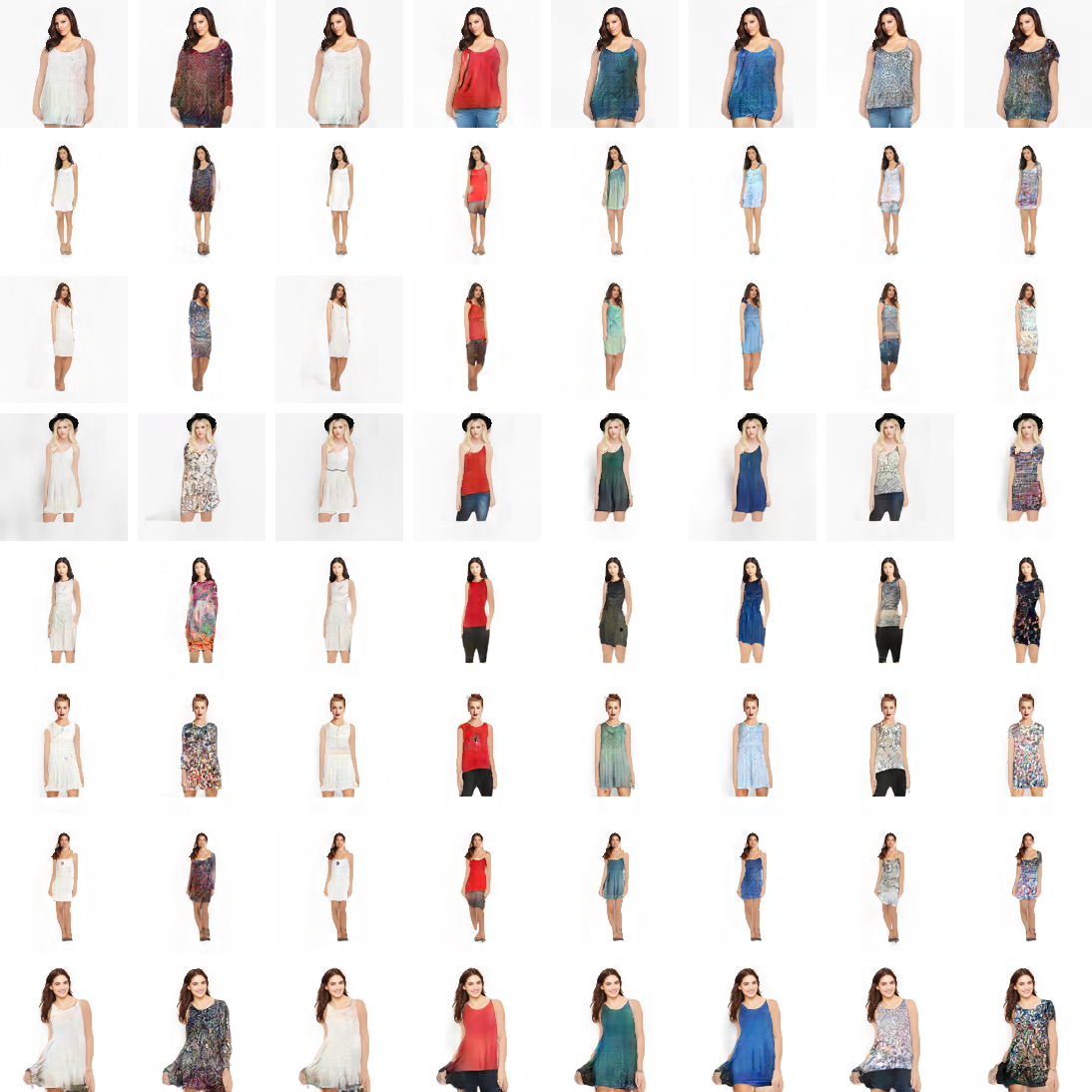}
\caption{\textbf{Matrix Visualization}. Each row of generated images is generated from the same original person while each column of generated images is conditioned on the same text description.}
\vspace{-0.25in}
\label{fig:matrix}
\end{center}
\end{figure*}

\noindent
\textbf{Matrix Visualization}. In this experiment, we visualize our results in an eight by eight matrix, where each row is generated by conditioning on the same original person, while each column is generated by conditioning on the same text description. We provide results in Fig.~\ref{fig:matrix}.

\begin{figure*}[t]
\begin{center}
\includegraphics[width=\linewidth]{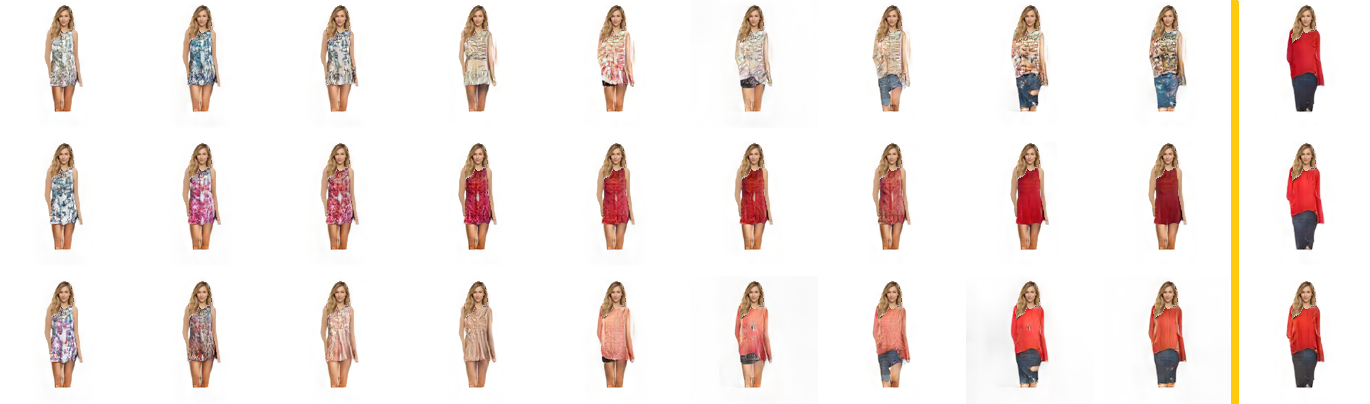}
\caption{\textbf{Walking the latent space}. For each row, the first and the last images are the two samples that we will make the interpolation. We gradually change the input from the left image. In the first row, we only interpolate the input to the first stage and hence the generated results only change in shapes. In the second row, we only interpolate the input to the second stage and hence the results only change in textures. The last row interpolate the input for both the first and second stages and hence the generated interpolated results transfer smoothly from the left to the right.}
\vspace{-0.25in}
\label{fig:interp}
\end{center}
\end{figure*}

\noindent
\textbf{Walking through the embedding space.} In this experiment, we generate the images by interpolating the embedding space (i.e. a concatenation of the input Gaussian noise and the text encoding), to show the gradual changes among the shapes and the textures of the generated clothes. We provide results in Fig.~\ref{fig:interp}. For each row, the first and the last images are the two samples that we will make the interpolation. We gradually change the input from the left image. In the first row, we only interpolate the input to the first stage and hence the generated results only change in shapes. In the second row, we only interpolate the input to the second stage and hence the results only change in textures. The last row interpolate the input for both the first and second stages and hence the generated interpolated results transfer smoothly from the left to the right.

\begin{figure}[t]
\begin{center}
\includegraphics[width=\linewidth]{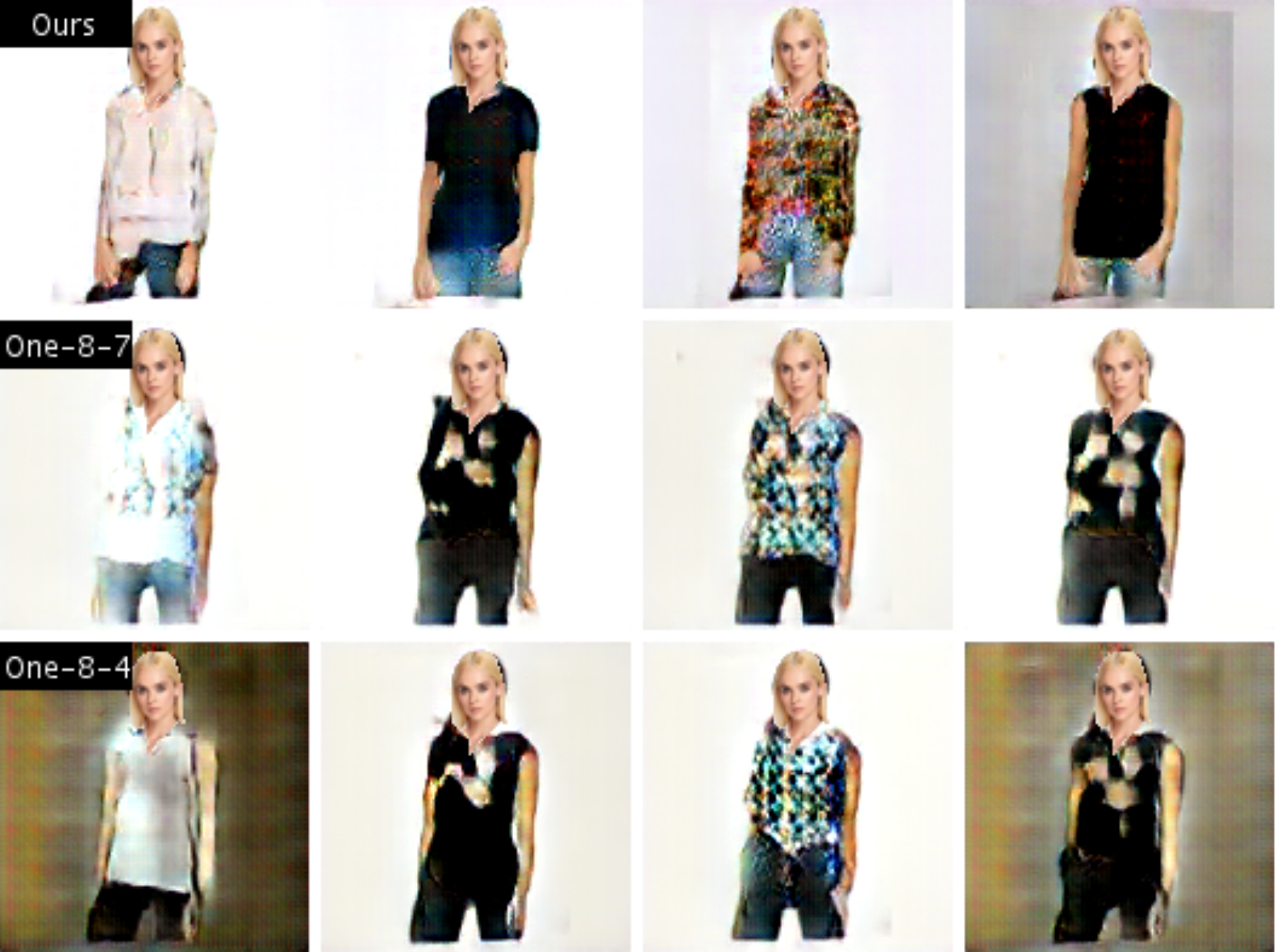}
\caption{\textbf{Comparison with one-step GAN baselines}. (We use abbreviation here. One-8-7 refers to One-Step-8-7 and One-8-4 refers to One-Step-8-4). Each row represents one approach and each column shares the same text input. While these baseline approaches can also keep the human shape to some extent, the artifact presented in their results are significantly stronger than our results. The shape of the generated persons are also less consistent for baseline approaches.}
\vspace{-0.25in}
\label{fig:e7}
\end{center}
\end{figure}

\noindent
\textbf{Comparison with One-Step GAN Baselines}. We provide a qualitative comparison with One-Step variants in Fig.~\ref{fig:e7}. As shown in the figure, our approach achieves better visual quality with fewer artifacts and more consistent human shape.

\begin{figure}[t]
\begin{center}
\includegraphics[width=\linewidth]{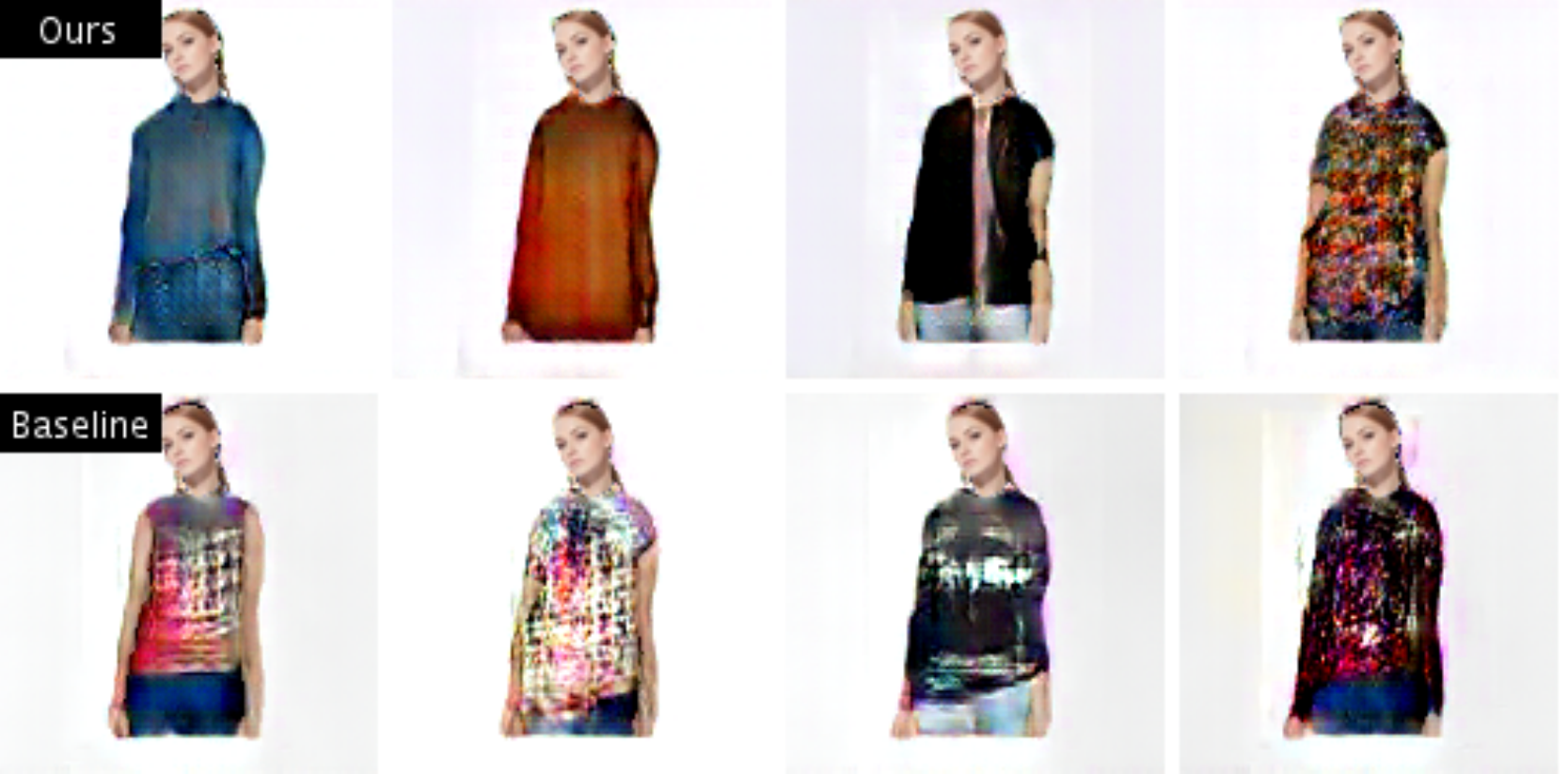}
\caption{\textbf{Comparison with the non-compositional baseline}. All the images are generated from the same human shape map from the previous stage. The upper row comes from the proposed compositional approach while the bottom row comes from the baseline method.}
\label{fig:e8}
\end{center}
\end{figure}

\noindent
\textbf{Comparison with the Non-Compositional Baseline}. We show the results in Fig.~\ref{fig:e8}. Our approach provides clearer clothing regions while much fewer visual artifacts and noise, outperforming the baseline approach.

\begin{figure}
\begin{center}
\includegraphics[width=\linewidth]{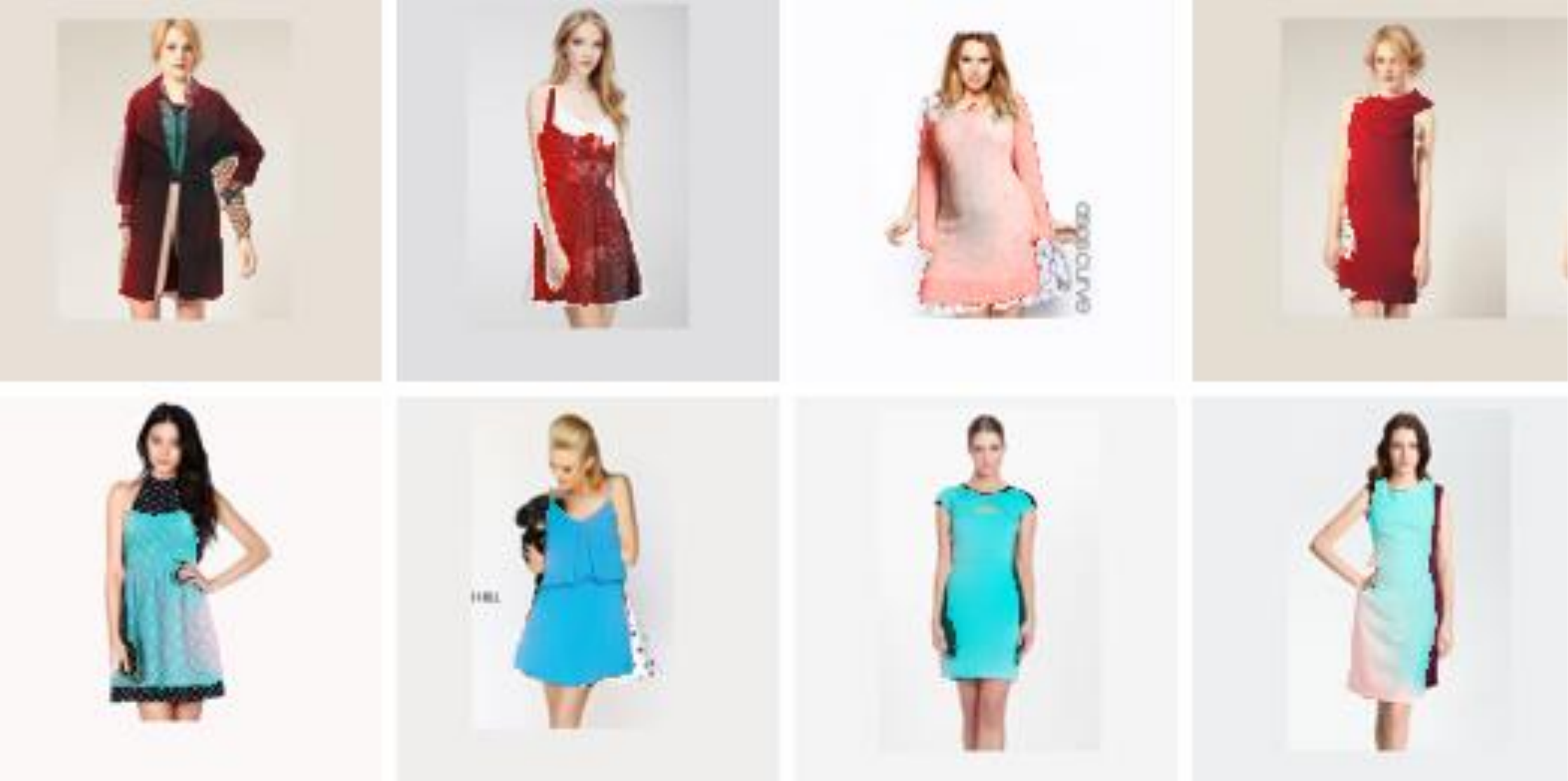}
\caption{\textbf{Representative failure cases for the 2D non-parametric baseline}. Shape inconsistency between the retrieved exemplar and body causes mismatch between the new outfit and the wearer.}
\vspace{-0.25in}
\label{fig:2d}
\end{center}
\end{figure}

\noindent
\textbf{Comparison with the 2D Non-Parametric Baseline}. We compare with this conventional baseline by retrieving an exemplar from a large database by text and perform Poisson image blending to apply the new outfit on the wearer. Results are shown in Fig.~\ref{fig:2d}. Due to shape inconsistency between the exemplar and wearer's body, the rendering results are not satisfactory.


\subsection{User Study}
\label{subsec:user_study}

\noindent
\textbf{Evaluating the Generated Segmentation Maps}.
A total of 50 volunteers participated in our user study. Our goal is to examine the quality of the intermediate segmentation maps generated by the first stage of FashionGAN.
To this end, we provided the human segmentation map of the original photograph and the generated map, \ie, a pair of maps for each test case, and asked participants to determine which map looked more realistic and genuine.
A higher number of misclassified test cases implies a better quality of the generated maps. As only FashionGAN would produce such intermediate segmentation map, we thus only conduct this experiment with our approach.
For the total of $8,979$ test cases, the participants misclassified $3,753$ of them (the misclassification rate is 42\%).
This is significant as our segmentation maps fooled most of the participants, whose ratings were close to random guessing.

\noindent
\textbf{Evaluating the Generated Photos.}
The same group of volunteers were asked to provide a ranking of the generated images produced by FashionGAN as well as the results from three baseline approaches, namely, `One-Step-8-7', `One-Step-8-4', and `Non-Compositional'. In addition, we also compared against the 2D non-parametric approach.
During the user study, each participant was provided with the original image and the corresponding sentence description.
The participants were asked to rank the quality of the generated images with respect to the relevance to the sentence description and the texture quality.

\begin{figure}
\begin{center}
\includegraphics[width=\linewidth]{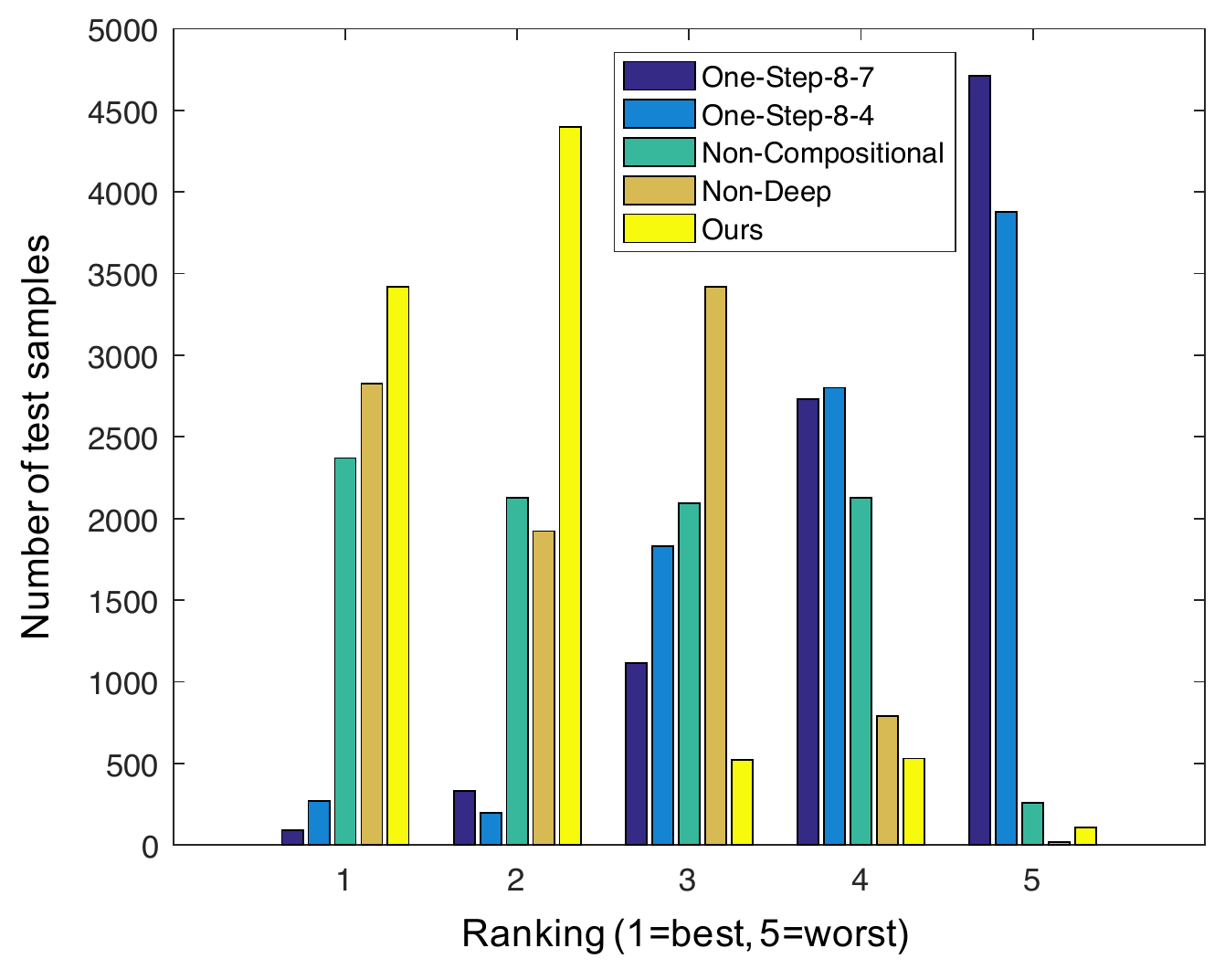}
\vskip -0.1cm
\caption{Detailed user study results. Each user rates 1 to 5 to each image they are assigned, and we show which methods each of these ratings go to. Rank 1 is the best and 5 is the worst.}
\vspace{-0.25in}
\label{fig:bar}
\end{center}
\end{figure}

All the 8,979 test images were evenly and randomly assigned to these participants.
We summarize various statistics in Table~\ref{tab:bar} and the frequency statistics for each rating in Fig.~\ref{fig:bar}. For each approach, we computed the average ranking (where 1 is the best and 5 is the worst), standard deviation, and the frequency of being assigned with each ranking.
We can observe that most of the high ranks go to our approach, which indicates that our solution achieves the best visual quality and relevance to the text input.

\begin{table}[t]
\centering
\caption{A user study that evaluates the quality of generated images. A smaller number indicates a higher rank.} \vspace{0.1cm}
\label{tab:bar}
\small{
\begin{tabular}{ccc}
\hline
                  & Mean Ranking & Std Ranking \\ \hline
One-Step-8-7      & 4.027       & 0.894      \\
One-Step-8-4      & 4.097       & 0.993      \\
Non-Compositional & 3.045       & 1.193      \\
2D Non-Parametric & 2.286       & 1.002      \\ \hline
Ours              & \textbf{1.544}       & \textbf{0.869}      \\ \hline
\end{tabular}
}
\end{table}

\section{Conclusion}
\label{sec:conclusion}
We presented a novel approach for generating new clothing on a wearer  based on textual descriptions. We designed two task-specific GANs, the shape and the image generators, and an effective spatial constraint in the shape generator. 
The generated images are shown to contain precise regions that are consistent with the description, while keeping the body shape and pose of a person unchanged. Quantitative and qualitative results outperform the baselines.
%

The results generated are limited by the current database we adopted. Our training set contains images mostly with a plain background as they were downloaded from on-line shopping sites (\ie,~\url{http://www.forever21.com/}). Hence the learning model is biased towards such a distribution.
In fact we do not assume any constraints or post-processing of the background. We believe that our model can also render textured background if the training set contains more images with textured background. The background distribution will be captured by the latent vector $\mathbf{z}$.

\vspace{0.1cm} \noindent
\textbf{Acknowledgement}: This work is supported by SenseTime Group Limited and the General Research Fund sponsored by the Research Grants Council of the Hong Kong SAR (CUHK 14224316, 14209217).

%

{\small
\bibliographystyle{ieee}
\bibliography{short,fashion}

\begin{thebibliography}{10}\itemsep=-1pt

\bibitem{gkioxari2015contextual}
G.~Gkioxari, R.~Girshick, and J.~Malik.
\newblock Contextual action recognition with {R*CNN}.
\newblock In {\em ICCV}, 2015.

\bibitem{goodfellow2014generative}
I.~Goodfellow, J.~Pouget-Abadie, M.~Mirza, B.~Xu, D.~Warde-Farley, S.~Ozair,
  A.~Courville, and Y.~Bengio.
\newblock Generative adversarial nets.
\newblock In {\em NIPS}, 2014.

\bibitem{isola2017image}
P.~Isola, J.-Y. Zhu, T.~Zhou, and A.~A. Efros.
\newblock Image-to-image translation with conditional adversarial networks.
\newblock In {\em CVPR}, 2017.

\bibitem{johnson2016perceptual}
J.~Johnson, A.~Alahi, and L.~Fei-Fei.
\newblock Perceptual losses for real-time style transfer and super-resolution.
\newblock In {\em European Conference on Computer Vision}, pages 694--711.
  Springer, 2016.

\bibitem{kingma2014adam}
D.~Kingma and J.~Ba.
\newblock Adam: A method for stochastic optimization.
\newblock {\em arXiv preprint arXiv:1412.6980}, 2014.

\bibitem{ledig2017photo}
C.~Ledig, L.~Theis, F.~Husz{\'a}r, J.~Caballero, A.~Cunningham, A.~Acosta,
  A.~Aitken, A.~Tejani, J.~Totz, Z.~Wang, et~al.
\newblock Photo-realistic single image super-resolution using a generative
  adversarial network.
\newblock In {\em CVPR}, 2017.

\bibitem{liang2015human}
X.~Liang, C.~Xu, X.~Shen, J.~Yang, S.~Liu, J.~Tang, L.~Lin, and S.~Yan.
\newblock Human parsing with contextualized convolutional neural network.
\newblock In {\em ICCV}, 2015.

\bibitem{liu2016deepfashion}
Z.~Liu, P.~Luo, S.~Qiu, X.~Wang, and X.~Tang.
\newblock {DeepFashion}: Powering robust clothes recognition and retrieval with
  rich annotations.
\newblock In {\em CVPR}, 2016.

\bibitem{nguyen2016synthesizing}
A.~Nguyen, A.~Dosovitskiy, J.~Yosinski, T.~Brox, and J.~Clune.
\newblock Synthesizing the preferred inputs for neurons in neural networks via
  deep generator networks.
\newblock In {\em NIPS}, 2016.

\bibitem{protopsaltou2002body}
D.~Protopsaltou, C.~Luible, M.~Arevalo, and N.~Magnenat-Thalmann.
\newblock A body and garment creation method for an internet based virtual
  fitting room.
\newblock In {\em Advances in Modelling, Animation and Rendering}, pages
  105--122. Springer, 2002.

\bibitem{radford2015unsupervised}
A.~Radford, L.~Metz, and S.~Chintala.
\newblock Unsupervised representation learning with deep convolutional
  generative adversarial networks.
\newblock {\em arXiv preprint arXiv:1511.06434}, 2015.

\bibitem{reed2016generative}
S.~Reed, Z.~Akata, X.~Yan, L.~Logeswaran, B.~Schiele, and H.~Lee.
\newblock Generative adversarial text-to-image synthesis.
\newblock In {\em ICMR}, 2016.

\bibitem{rother2004grabcut}
C.~Rother, V.~Kolmogorov, and A.~Blake.
\newblock Grabcut: Interactive foreground extraction using iterated graph cuts.
\newblock {\em TOG}, 23(3):309--314, 2004.

\bibitem{salimans2016improved}
T.~Salimans, I.~Goodfellow, W.~Zaremba, V.~Cheung, A.~Radford, and X.~Chen.
\newblock Improved techniques for training {GANs}.
\newblock In {\em NIPS}, 2016.

\bibitem{wang2016generative}
X.~Wang and A.~Gupta.
\newblock Generative image modeling using style and structure adversarial
  networks.
\newblock In {\em ECCV}, 2016.

\bibitem{yang2016detailed}
S.~Yang, T.~Ambert, Z.~Pan, K.~Wang, L.~Yu, T.~Berg, and M.~C. Lin.
\newblock Detailed garment recovery from a single-view image.
\newblock {\em arXiv preprint arXiv:1608.01250}, 2016.

\bibitem{yoo2016pixel}
D.~Yoo, N.~Kim, S.~Park, A.~S. Paek, and I.~S. Kweon.
\newblock Pixel-level domain transfer.
\newblock In {\em ECCV}, 2016.

\bibitem{zhang2017stackgan}
H.~Zhang, T.~Xu, H.~Li, S.~Zhang, X.~Huang, X.~Wang, and D.~Metaxas.
\newblock Stackgan: Text to photo-realistic image synthesis with stacked
  generative adversarial networks.
\newblock In {\em ICCV}, 2017.

\bibitem{zhou2010parametric}
S.~Zhou, H.~Fu, L.~Liu, D.~Cohen-Or, and X.~Han.
\newblock Parametric reshaping of human bodies in images.
\newblock {\em TOG}, 29(4):126, 2010.

\end{thebibliography}
}

\newpage

\end{document}